%% file: iclr2022_conference.tex
\documentclass{article} 
\usepackage{iclr2022_conference,times}

\input{math_commands.tex}

\usepackage{hyperref}
\usepackage{url}
\usepackage{multirow}
\usepackage[ruled]{algorithm2e}
\usepackage{amsthm,amsmath,amssymb}
\usepackage{mathrsfs}
\usepackage{graphicx}
\usepackage{threeparttable}
\usepackage{ amssymb }

\title{DropAttack: A Masked Weight Adversarial Training Method to Improve Generalization of Neural Networks}

\author{Shiwen Ni, Jiawen Li \& Hung-Yu Kao\thanks{Corresponding author} \\
Department of Computer Science and Information Engineering\\
National Cheng Kung University\\
Tainan, Taiwan \\
\texttt{\{P78083033, P78073012\}@gs.ncku.edu.tw, hykao@mail.ncku.edu.tw} \\
}

\iclrfinalcopy 
\begin{document}

\maketitle

\begin{abstract}
Adversarial training has been proven to be a powerful regularization method to improve the generalization of models. However, current adversarial training methods only attack the original input sample or the embedding vectors, and their attacks lack coverage and diversity. To further enhance the breadth and depth of attack, we propose a novel masked weight adversarial training method called DropAttack, which enhances generalization of model by adding intentionally worst-case adversarial perturbations to both the input and hidden layers in different dimensions and minimize the adversarial risks generated by each layer. DropAttack is a general technique and can be adopt to a wide variety of neural networks with different architectures. To validate the effectiveness of the proposed method, we used five public datasets in the fields of natural language processing (NLP) and computer vision (CV) for experimental evaluating. We compare the proposed method with other adversarial training methods and regularization methods, and our method achieves state-of-the-art on all datasets. In addition, Dropattack can achieve the same performance when it use only a half training data compared to other standard training method. Theoretical analysis reveals that DropAttack can perform gradient regularization at random on some of the input and wight parameters of the model. Further visualization experiments show that DropAttack can push the minimum risk of the model to a lower and flatter loss landscapes. Our source code is publicly available on github\footnote{https://github.com/nishiwen1214/DropAttack}.
\end{abstract}

\section{Introduction}

Deep neural networks (DNNs) \citep{lecun2015deep} have achieved state-of-the-art performance in many artificial intelligence applications, such as natural language processing and computer vision. Regularization methods such as L1 \cite{tibshirani1996regression} and L2 \citep{tikhonov1943stability} regularization, early stopping \citep{morgan1989generalization} and Dropout \citep{srivastava2014dropout}, play an important role in the impressive performance of deep networks by controlling the model complexity, and thus preventing overfitting and improving generalization. Adversarial training \citep{goodfellow2014explaining} was originally proposed as a method to improve the security of machine learning systems in order to train a neural network that is robust to attack samples. And adversarial training is the process of training a model, which minimizes the maximal risk for label-preserving input perturbations. It improves not only robustness to adversarial examples, but also generalization performance for original examples. \citet{goodfellow2014explaining}  have demonstrated that adversarial training can result in regularization; even further regularization than dropout.

In this work, we mainly focus on improving the generalization performance of the model and preventing the model from overfitting, rather than enhancing the robustness of model to attack samples. \citet{miyato2016adversarial} applies adversarial training to text classification tasks, and finds that adversarial training can effectively improve the generalization of text or RNN models on the test set. 
Most of the recent adversarial training is aimed at the input of the model to attack. Moreover, these existing adversarial training methods add perturbation to every element in the input tensor during the attack. (It should be noted that the input in the NLP field is the embeddings of the text, and in the CV field it is the value of each pixel of the picture. In this paper, it is uniformly called the input for the convenience of expression.) In order to increase the breadth of the attack, we expanded the attack target from the input to the weight parameters of other layers, that is, in the process of adversarial training, while attacking the input, it also attacks the weight parameters of other layers. In each iteration of the attack, we randomly mask the attack on a certain proportion elements instead of attacking all the elements in the input or weight tensor. In this way, exponentially different attack combinations can be obtained, and the internal adversarial loss of the model can be maximized. 

In this paper, we show the impact of DropAttack and various other well-known regularization methods on the generalization performance of the model. We experiment with different neural network models on five different public datasets to prove the effectiveness of the proposed method. We visually analyze the training and verification accuracy of some models of different architectures as the training progresses. And we also analyzed the impact of hyperparameters and optimization of multi-forward-backward propagation on DropAttack. Finally, we also conducted a theoretical analysis of the proposed method from another perspective to prove the effectiveness of DropAttack.
\section{Related work}
\label{2}
Adversarial training can be traced back to \citep{goodfellow2014explaining}, in which the model improves the robustness and generalization of the model by generating adversarial examples and injecting them into training data. The effectiveness of adversarial training largely depends on the direction of the attack, so it is necessary to find the perturbation value that maximizes the adversarial loss. Due to its linear characteristics, neural networks are easily attacked by linear perturbation. Therefore, \citet{goodfellow2014explaining} proposed the Fast Gradient Sign Method (FGSM) to calculate the perturbation of the input sample. They linearized the cost function around the current value of parameters, obtaining an optimal max-norm constrained pertubation of:
\begin{equation}\label{key}
\bm{r_{adv}} =\epsilon \cdot {\rm sgn}( \nabla_{\bm{x}}L(\bm{\theta},\bm{x},y))
\end{equation}
Where $\bm{\theta}$ is the model parameter, $\bm{x}$ is the input of the model, y is the label corresponding to the input, $L$ is the cost used to train the neural network. sgn is the symbolic function, and $\epsilon$ is the perturbation coefficient. In order to find a better perturbation, \citet{miyato2016adversarial} proposed the Fast Gradient Method (FGM), which made a simple modification to the calculation of perturbation in FGSM. And FGM is the first time that adversarial training has been applied to text classification tasks. The formula is as follows:
\begin{equation}\label{key}
\bm{r_{adv}}= \epsilon \cdot \bm{g} /\Vert\bm{g}\Vert_2  {~\rm where~} \bm{g}=\nabla_{\bm{x}}L(\bm{\theta},\bm{x},y).
\end{equation}

\citet{athalye2018obfuscated} proposed a confrontation training method called Projected Gradient Descent (PGD), which obtains the final perturbation value through multiple forward and backward propagation iterations, and the perturbation obtained in each iteration is limited to a set range, if it exceeds this range will be mapped to the “sphere” of the range. To put it simply, “walk in small steps, take a few more steps.” The formula is as follows:
\begin{equation}\label{key}
x_{t-1}=\prod_{x+S}(x_t +\alpha \cdot \bm{g(x_t)} /\Vert\bm{g(x_t)}\Vert_2 ) {~\rm where~} \bm{g(x_t)}=\nabla_{\bm{x}}L(\bm{\theta},\bm{x_t},y).
\end{equation}
where $S=r\in \mathbb{R}^d ; \Vert\bm{r}\Vert_2 \leq \epsilon$ is the constraint space of the perturbation, and a is the step length of the “small step”.

After that, although many defenses were broken by \citet{athalye2018obfuscated}, PGD-based adversarial training is one of the few that can withstand powerful attacks. \citet{athalye2018obfuscated} proved that PGD can largely avoid the problem of gradient confusion, but it will still cause high convolution and non-linear loss surface. When K is very small, it will be lightly broken under powerful adversaries. To get more efficient PGD-based adversarial training, it must iteratively calculate the gradient many times, but this will consume a lot of computing resources. \citet{shafahi2019adversarial} proposed a “free” adversarial training algorithm that eliminates the overhead cost of calculating adversarial perturbations by recycling the gradient information computed when updating model parameters. \citet{zhang2019you} effectively reduce the total number of full forward and backward propagations by restricting most of the forward and back propagation within the first layer of the network during adversary updates. \citet{miyato2019virtual} proposed virtual adversarial training as a regularization method for semi-supervised learning in the text field. \citet{zhu2020freelb} propose FreeLB for improving the generalization of language models, which performs multiple PGD iterations to attack the embeddings, and simultaneously accumulates the “free” parameter gradients in each iteration.

In this work, we are the first to propose an adversarial training method that simultaneously attacks the input of the model and the weight parameters of other layers to improve the generalization of the model. And our method, DropAttack, is the first to use random masking of some elements to increase the diversity of adversarial attack combinations.

\section{The Proposed DropAttack Adversarial Training Method}
\label{3}

\begin{figure*}[h]
	\centering
	\includegraphics[width=0.9\linewidth]{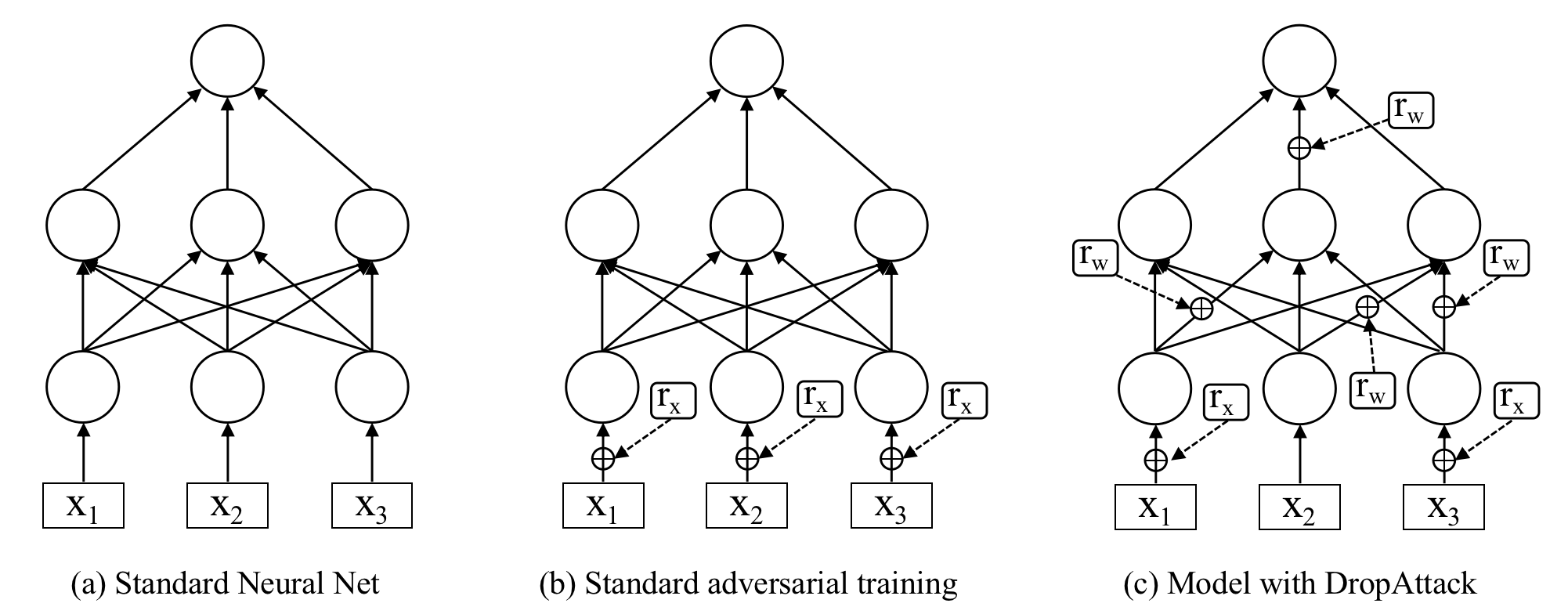}
	\caption{Neural network model (a) with standard adversarial Training (b) and with DropAttack (c). (a) A standard neural network with 3 input and 12 weight parameters. (b) An new neural network produced by applying standard adversarial Training on the left, which attacks all inputs. (c) An new neural network produced by applying DropAttack to the network on the left. Assume that each input vecter and weight parameter have a 2/3 and 1/3 probability of being attacked , respectively.
	}
	\label{f1}
\end{figure*}

The proposed new adversarial training method is inspired by Dropout, so we named it DropAttack. Standard adversarial training is to explore the optimal parameters to minimize the maximum risk of adversarial attacks. The Min-Max formula is as follows:
\begin{equation}\label{key}
\mathop{{\rm min}}\limits_{\bm{\theta}} \mathbb{E}_{(x,y)\sim D}[\mathop{{\rm max}}\limits_{r_{adv}\in S}L(\bm{\theta},\bm{x}+r_{adv},y) ]
\end{equation}
where $D$ is the data distribution, $y$ is the label, and L is loss function. $r_{adv}$ is the perturbation under maximizing internal risk. $S$ is the perturbation constraint space. Here we propose a new adversarial training method, DropAttack, which simultaneously attacks the input x of the model and the weight parameters of other layers, and randomly masks some attacks.  The overall procedure is shown in Algorithm \ref{a1}. The Min-Max formula of DropAttack can be expressed as:

\begin{equation}\label{key}
\mathop{{\rm min}}\limits_{\bm{\theta}} \mathbb{E}_{(x,y)\sim D}[\mathop{{\rm max}}\limits_{r_{x}\in S}L(\bm{\theta},\bm{x}+\bm{M_{x}}\cdot \bm{r_x},y)+\mathop{{\rm max}}\limits_{r_{{\theta}}\in S}L(\bm{\theta}+\bm{M_{\theta}}\cdot \bm{r_\theta},\bm{x},y) ]
\end{equation}
where $\bm{r_x}$ and $\bm{r_\theta}$ are the perturbation of the input x and the parameter $\bm{\theta}$ under maximizing the internal risk. We respectively approximate these values by linearizing  $\nabla_{\bm{x}}L(\bm{\theta},\bm{x_t},y)$ and $\nabla_{\bm{{\theta}}}L(\bm{\theta},\bm{x_t},y)$ around $\bm{x}$ and $\bm{\theta}$. Using the linear approximation in equation (6) and the L2 norm constraint, the resulting adversarial perturbation is
\begin{equation}\label{key}
\bm{r_x}\leftarrow \epsilon_x \cdot \nabla_{\bm{x}}L(\bm{\theta},\bm{x},y) /\Vert\nabla_{\bm{x}}L(\bm{\theta},\bm{x},y)\Vert_2  ;~
\bm{r_{\theta}}\leftarrow \epsilon_{\theta} \cdot \nabla_{\bm{{\theta}}}L(\bm{\theta},\bm{x},y) /\Vert\nabla_{\bm{{\theta}}}L(\bm{\theta},\bm{x},y)\Vert_2
\end{equation}
These perturbation can be easily calculated using backpropagation in a neural network.

$\bm{M_{x}}$ and $\bm{M_{\theta}}$ are the Random attack masks of $r_x$ and $r_\theta$ respectively. For any attack mask, $\bm{M_{ij}}$ is a matrix of independent Bernoulli random variables with the same dimension as the perturbation value, and the probability of each Bernoulli random variable is 1. Multiplying the perturbation matrix and the attack mask matrix will randomly mask a part of the element values in the perturbation matrix. 

\begin{algorithm}[t]
	\label{a1}
	\caption{DropAttack Adversarial Training}\label{algorithm}
	\KwIn{Training samples $\mathcal{X}$ , model parameter $\bm{\theta}$, perturbation coefficients $\epsilon_x$ and $\epsilon_\theta$, Attack probabilities $p_x$ and $p_\theta$  , Learning rate $\tau$}
	
	\For{${\rm epoch} = 1 \ldots N_{ep}$}{\For{$(\bm{x},y)\in \mathcal{X}$}{{
				Compute gradient $\bm{g}$ of parameter $\bm{x}$ and $\bm{\theta}$:\\
				\hspace{2em}$\bm{g_x} \leftarrow \nabla_{\bm{x}}L(\bm{\theta},\bm{x},y)$; $\bm{g_{\theta}} \leftarrow  \nabla_{\bm{\theta}}L(\bm{\theta},\bm{x},y)$
				
				Compute perturbation $\bm{r_x}$ and $\bm{r_{\theta}}$:\\
				\hspace{2em}$\bm{r_x}\leftarrow \epsilon_x \cdot \bm{g_x} /\Vert\bm{g_x}\Vert_2$; $\bm{r_{\theta}}\leftarrow \epsilon_{\theta} \cdot \bm{g_{\theta}} /\Vert\bm{g_{\theta}}\Vert_2$
				
				Random attack $\bm{M_{x}}$ and $\bm{M_{{\theta}}}$ mask:\\
				\hspace{2em}$\bm{M_{{x}_{ij}}}\sim {\rm Bernoulli}(p_x)$; $\bm{M_{{{\theta}}_{ij}}}\sim {\rm Bernoulli}(p_{\theta})$
				
				Compute adversarial gradient $\bm{g}_{adv}$:\\
				\hspace{2em}$\bm{g}_{adv}\leftarrow \nabla_{\bm{\theta}}[L(\bm{\theta},\bm{x}+\bm{M_{x}}\cdot \bm{r_x},y)+L(\bm{\theta}+\bm{M_{\theta}}\cdot \bm{r_\theta},\bm{x},y)]$
				
				Update parameter $\bm{\theta}$: \\
				\hspace{2em}$\bm{\theta} \leftarrow \bm{\theta}- \tau(\bm{g}+\bm{g}_{adv})$}}}
	\KwOut{\bm{$\theta$}}
\end{algorithm}

The attack on the input in the text model is essentially an attack on the weight parameters of the embedding layer. We can see that Figure \ref{f1} (a) is a standard neural network with 12 weight parameters. Figure \ref{f1} (b) is an new neural network produced by applying standard adversarial Training on the left, which attacks all inputs. And Figure \ref{f1} (c) is an new neural network after DropAttack is applied, and some of the weights are added with the perturbation calculated by the corresponding gradient. Compared with standard adversarial training, DropAttack maximizes internal risk through a wider range of attacks (not limited to the input layer). And randomly mask some dimensional perturbation, which will generate a more robust and diversified embedding space.

\subsection{DropAttack with multiple internal ascent steps}

Most of the latest adversarial training are PGD-based methods. PGD-based methods are a series of adversarial training algorithms \citep{kurakin2016adversarial} for solving the maximum-minimum problem of cross-entropy loss, which can be reliably achieved by using multiple projection gradient ascent steps and then performing an SGD (Stochastic Gradient Descent) step. Multiple PGD iterations can get a more optimized perturbation. Obviously, DropAttack can also use multiple forward and backward propagation methods to update the perturbation. In practice, when DropAttack is used, each ascent step of updating perturbation is optimizing for a different weight network. Because $\bm{r_t}$ and $\bm{r_{t-1}}$ are an iterative relationship, this will affect the inability to obtain the optimal perturbation. Therefore, we use the same mask matrix for each step of perturbation update. The overall procedure is shown in Algorithm \ref{a2}. We first calculate the initial gradients $\bm{g^{(0)}_x} = \nabla_{\bm{x}}L(\bm{\theta},\bm{x},y)$ and   $\bm{g^{(0)}_{\theta}} =  \nabla_{\bm{\theta}}L(\bm{\theta},\bm{x},y)$  of $\bm{x}$ and $\bm{\theta}$, and initial perturbation $\bm{r^{(0)}_x}= \epsilon_x \cdot \bm{g^{(0)}_x} /\Vert\bm{g^{(0)}_x}\Vert_2$,   $\bm{r^{(0)}_{\theta}}= \epsilon_{\theta} \cdot \bm{g^{(0)}_{\theta}} /\Vert\bm{g^{(0)}_{\theta}}\Vert_2$. Then generate random attack masks $\bm{M_{x}}$ and $\bm{M_{{\theta}}}$, which are used in the forward and backward propagation for each step of perturbation update. That is, the mask matrix is fixed after the first step of perturbation update. The perturbation values $\bm{r^{(t)}_x} = \epsilon_x \cdot \bm{g^{(t)}_x} /\Vert\bm{g^{(t)}_x}\Vert_2$ and  $\bm{r^{(t)}_{\theta}}= \epsilon_{\theta} \cdot \bm{g^{(t)}_{\theta}} /\Vert\bm{g^{(t)}_{\theta}}\Vert_2$ are updated through the gradient ascend in each iteration, where $\bm{g^{(t)}_x}= \bm{g^{(t-1)}_x} + \frac{1}{K} \nabla_{\bm{x}}L(\bm{\theta},\bm{x}+\bm{M_{x}}\cdot \bm{r^{(t-1)}_x},y)$ and $\bm{g^{(t)}_{\theta}}= \bm{g^{(t-1)}_{\theta}} + \frac{1}{K} \nabla_{\bm{\theta}}L(\bm{\theta},\bm{x}+\bm{M_{{\theta}}}\cdot \bm{r^{(t-1)}_{\theta}},y)$.
Finally, the model parameter $\bm{\theta}$ is updated once with the accumulated gradient of each adversarial iteration, the DropAttack-K of K iterations can be expressed as:
\begin{equation}\label{key}
\mathop{{\rm min}}\limits_{\bm{\theta}} \mathbb{E}_{(x,y)\sim D}\{\dfrac{1}{K}\sum^{K-1}_{t=0}[\mathop{{\rm max}}\limits_{r^{(t)}_{x}\in S}L(\bm{\theta},\bm{x}+\bm{M_{x}}\cdot \bm{r^{(t)}_x},y)+\mathop{{\rm max}}\limits_{r^{(t)}_{{\theta}}\in S}L(\bm{\theta}+\bm{M_{\theta}}\cdot \bm{r^{(t)}_\theta},\bm{x},y) ]\}.
\end{equation}

\begin{algorithm}[h]
	\caption{PGD-based DropAttack-K Adversarial Training}\label{a2}
	\KwIn{Training samples $\mathcal{X}$ , model parameter $\bm{\theta}$, perturbation coefficient $\epsilon_x$ and $\epsilon_\theta$, Attack probability $p_x$ and $p_\theta$, number of forward-backward propagation K, Learning rate $\tau$}
	
	\For{${\rm epoch} = 1 \ldots N_{ep}$}{\For{$(\bm{x},y)\in \mathcal{X}$}{\For{$\bm{t}=1,2\ldots K$}{
				
				Compute initial gradient $\bm{g}$ of parameter $\bm{x}$ and $\bm{\theta}$:\\
				\hspace{2em}$\bm{g^{(0)}_x} \leftarrow \nabla_{\bm{x}}L(\bm{\theta},\bm{x},y)$; $\bm{g^{(0)}_{\theta}} \leftarrow  \nabla_{\bm{\theta}}L(\bm{\theta},\bm{x},y)$

				Compute initial perturbation $\bm{r^{(0)}_x}$ and $\bm{r^{(0)}_{\theta}}$:\\
				\hspace{2em}$\bm{r^{(0)}_x}\leftarrow \epsilon_x \cdot \bm{g^{(0)}_x} /\Vert\bm{g^{(0)}_x}\Vert_2$; $\bm{r^{(0)}_{\theta}}\leftarrow \epsilon_{\theta} \cdot \bm{g^{(0)}_{\theta}} /\Vert\bm{g^{(0)}_{\theta}}\Vert_2$
				
				\If{$\bm{t}=1$}{Generate random attack masks $\bm{M_{x}}$ and $\bm{M_{{\theta}}}$:\\
					\hspace{2em}$\bm{M_{{x}_{ij}}}\sim {\rm Bernoulli}(p_x)$; $\bm{M_{{{\theta}}_{ij}}}\sim {\rm Bernoulli}(p_{\theta})$}
				
				Update the perturbation $\bm{r}$ via gradient ascend :\\
				\hspace{2em}$\bm{g^{(t)}_x}\leftarrow \bm{g^{(t-1)}_x} + \frac{1}{K} \nabla_{\bm{x}}L(\bm{\theta},\bm{x}+\bm{M_{x}}\cdot \bm{r^{(t-1)}_x},y)$
				
				\hspace{2em}$\bm{g^{(t)}_{\theta}}\leftarrow \bm{g^{(t-1)}_{\theta}} + \frac{1}{K} \nabla_{\bm{\theta}}L(\bm{\theta},\bm{x}+\bm{M_{{\theta}}}\cdot \bm{r^{(t-1)}_{\theta}},y)$
				
				\hspace{2em}$\bm{r^{(t)}_x}\leftarrow \epsilon_x \cdot \bm{g^{(t)}_x} /\Vert\bm{g^{(t)}_x}\Vert_2$; $\bm{r^{(t)}_{\theta}}\leftarrow \epsilon_{\theta} \cdot \bm{g^{(t)}_{\theta}} /\Vert\bm{g^{(t)}_{\theta}}\Vert_2$
			}
			Update parameter $\bm{\theta}$: \\
			\hspace{2em}$\bm{\theta} \leftarrow \bm{\theta}- \tau(\bm{g^{(K)}_x}+\bm{g^{(K)}_{\theta}})$}}
	\KwOut{\bm{$\theta$}}
\end{algorithm}

The training process is equivalent to replacing the original batch with a K-times virtual batch, consisting of samples whose embeddings are $\bm{x}+\bm{M_{x}}\cdot \bm{r^{(0)}_x}, \bm{x}+\bm{M_{x}}\cdot \bm{r^{(1)}_x},\ldots ,\bm{x}+\bm{M_{x}}\cdot \bm{r^{(k-1)}_x}$. Similarly, multiple virtual neural networks with different weight parameters will be trained, and their weight parameters are $\bm{{\theta}}+\bm{M_{{\theta}}}\cdot \bm{r^{(0)}_{\theta}}, \bm{{\theta}}+\bm{M_{{\theta}}}\cdot \bm{r^{(1)}_{\theta}},\ldots ,\bm{{\theta}}+\bm{M_{{\theta}}}\cdot \bm{r^{(k-1)}_{\theta}}$ respectively. It is worth noting that our perturbation constraint $S$ does not take an additional fixed value, and only uses the L2 norm to constrain the gradient value, because we need the diversity of perturbations at each step, rather than forcibly constraining it in a fixed spherical space. In fact, our DropAttack-K also inherits the "free" ability, using the gradient average calculated by each backpropagation for external minimization.

Intuitively, compared to the previous adversarial training method, DropAttack can generate a richer adversarial sample in the spherical space of the original sample, which can prevent the model from overfitting on the adversarial sample to a certain extent. Empirically, there is a certain gap between the features of the training dataset and the features of the test dataset in the high-dimensional feature space. Improving generalization is essentially to narrow the feature distribution gap between the training dataset and the test dataset. However, this gap is uncertain, so more diverse adversarial samples are needed to fill this gap. In theory, DropAttack has a more significant improvement in the generalization of the model.

\section{Experiment}
\label{4}
In this section, we test and analyze the effect of DropAttack on three NLP data sets and two CV data sets. In addition, we also analyze the ability of DropAttack to prevent overfitting under different sizes of training data. Additional experimental details and results are provided in the Appendix \ref{a}.
\subsection{Datasets}
Five public datasets, IMDB \citep{maas2011learning}, PHEME \citep{zubiaga2016learning}, AGnews \citep{zhang2015character}, MNIST \citep{lecun1998mnist} and GIFAR-10 \citep{krizhevsky2009learning}, are used to evaluate our DropAttack algorithm. The brief description of the datasets are shown in Table \ref{t1}. IMDB (Maas et al., 2011) is a standard benchmark movie review dataset for sentiment analysis. PHEME dataset contains a collection of Twitter rumours and non-rumours posted during breaking news. The AGnews topic classification dataset is constructed by Xiang Zhang from the original AG news sources. MNIST is s standard and commonly used toy dataset of handwritten digits. The CIFAR-10 dataset consists of 60000 32x32 colour images in 10 classes, with 6000 images per class.  We divide each dataset into training set, validation set and test set.
\begin{table}[h]
	\caption{Overview of the datasets used in this paper.}
	\label{t1}
	\centering
	\begin{tabular}{l|l|c|c|c|c}
		\hline
		Dataset & Task & \multicolumn{1}{l|}{Classes} & Training & Validation & Test \\ \hline \hline
		IMDB & Sentiment analysis & 2 & 40000 & 5000 & 5000 \\
		PHEME & Rumor detection & 2 & 5145 & 643 & 637 \\
		AGnews & News classification & 4 & 110000 & 10000 & 7600 \\
		MNIST & Image classification & 10 & 50000 & 10000 & 10000 \\
		CIFAR-10 & Image classification & 10 & 40000 & 10000 & 10000 \\ \hline
	\end{tabular}
\end{table}
\subsection{Experimental Setup}
In the experiment, we chose the rnn-based model and the cnn-based model to handle nlp tasks and cv tasks, respectively. For nlp tasks, IMDB uses an LSTM \citep{hochreiter1997long} (300-300 dim) layer and a fully connected layer (300-2 dim); PHEME uses a BiGRU \citep{cho2014learning} (300-300 dim) layer and a fully connected layer (600-2 dim); AGnews uses two BiLSTM \citep{schuster1997bidirectional} (300-300 dim) layers and a fully connected layer (600-4 dim). And for cv tasks, MNIST uses the LeNet-5 \citep{lecun1998gradient} model, which contains two layers of CNN(1-6-16 channels, kernel size = 5) and three fully connected layers (400-120-84-10 dim); CIFAR-10 uses the VGGNet-16 \citep{simonyan2014very} model, which ontains 13 layers of CNN (3-64-64-128-128-256-256-256-512-512-512-512 channels, kernel size = 3) and three fully connected layers (512-4096-4096-10 dim). And all models are implemented based on Pytorch, the Batch sizes value is 128, the optimizer is Adam, and the learning rate is 0.001. 

\textbf{Methods used for comparison}:

\textbf{L1} \citep{tibshirani1996regression} \textbf{and L2} \citep{tikhonov1943stability} \textbf{regularization}: Regular constraints are added to the original objective function, the L1 norm conforms to the Laplace distribution, and the L2 norm conforms to the Gaussian distribution.

\textbf{Dropout} \citep{srivastava2014dropout}: It is a commonly used regularization method that randomly removes some neural units and all their input and output connections during the training process. It prevents overfitting and provides a way of approximately combining exponentially many different neural network architectures effciently.

\textbf{FGSM} \citep{goodfellow2014explaining}: The first adversarial training method uses the sign function to generate the perturbation based on the direction of the gradient ascend.

\textbf{FGM} \citet{miyato2016adversarial}: Compared with the FGSM method, FGM improves the calculation method of perturbation. And the perturbation is obtained by dividing the linear approximation of the gradient by the L2 regularization of the gradient.

\textbf{PGD} \citet{athalye2018obfuscated}: An adversarial training method that requires K forward-backward passes through the network to calculate the optimal perturbation.

\textbf{FreeAT} \citet{shafahi2019adversarial} : A PGD-baesd adversarial training method, which accelerates the training process by sharing the gradient of internal maximization and external minimization.

\textbf{FreeLB} \citet{zhu2020freelb}: A PGD-baesd adversarial training method, which uses the gradient average calculated by K steps to update the model parameters.
\subsection{Experimental Results and Discussion}
\begin{table}[]
	\caption{Comparing the effects of DropAttack with various well-known regularization and other state-of-the-art adversarial training methods. The reported results are calculated from 5 runs with the same hyper-parameters except for the random seeds.  The "Original models" used in the five datasets are LSTM, BiGRU, BiLSTM, LeNet and VGGNet, which are common neural-based deep learning models. DropAttack-(I) and DropAttack-(I\&W) are the attack input only and the simultaneous attack input and weight parameters, respectively. The best method and the best competitor are highlighted by \textbf{bold} and \underline{underline}, respectively. }
	\label{t2}
	\begin{center}
		\begin{threeparttable}
		\begin{tabular}{l|c|c|c|c|c}
			\hline
			\multirow{2}{*}{Methods} & \multicolumn{3}{c|}{NLP Datasets} & \multicolumn{2}{c}{CV Datasets} \\ \cline{2-6} 
			& IMDB & PHEME\tnote{*}& AGnews & MNIST & CIFAR-10 \\ \hline \hline
			Original model & 88.12 & 84.08/78.99 & 91.87 & 98.95 & 84.67 \\ \hline \hline
			Original model + L1 & 88.02 & 85.34/79.55 & 92.29 & 99.07 & 84.74 \\
			Original model + L2 & 88.27 & 85.67/\underline{81.29} & 92.43 & 99.14 & 84.63 \\
			Original model + Dropout & 88.64 & 85.85/81.08 & 92.22 & 99.07 & 85.39 \\ \hline 
			Original model + FGSM & 88.04 & 85.61/80.40 & 92.54 & 99.16 & 71.86 \\
			Original model + FGM & 89.26 & 84.97/78.52 & 92.53 & \underline{99.15} & \underline{85.64} \\
			Original model + PGD & \underline{89.38} & 85.28/79.30 & \underline{92.76} & 99.10 & 85.57 \\
			Original model + FreeAT & 89.17 & 85.29/79.32 & 92.45 & 99.09 & 85.45 \\
			Original model + FreeLB & 89.25 & \underline{85.69}/81.18 & 92.58 & 99.11 & 85.47 \\ \hline
			Original model + DropAttack-(I) & 89.76	& 85.75/81.33 & 93.35 & 99.16 & 86.05 \\ 
			Original model + DropAttack-(I\&W) & \textbf{90.36} & \textbf{87.15/81.31} &\textbf{93.37} & \textbf{99.27}& \textbf{86.09} \\ \hline
		\end{tabular}
	\begin{tablenotes}
		\footnotesize
		\item[*] {Note that although PHEME is a two-class classification, the labels are not balanced, so we use accuracy and f1-score (accuracy/f1) as the evaluation criteria.} 
	\end{tablenotes}
\end{threeparttable}
	\end{center}
\end{table}

We can find from Table \ref{t2} that compared to the original model without DropAttack, the performance of the model after using DropAttack has improved on the five datasets, and the improvements on the three nlp datasets are 2.24\%, 3.07\% and 1.5\% respectively. Compared with other regularization methods, the overall effect of adversarial training is better. Among them, the efficiency of FGSM is relatively unstable, and the performance on IMDB and CIFAR-10 is only 88.04\% and 71.86\% respectively. Because the naive perturbation value may destroy the distribution of the original data. It should be noted that PGD, FreeAT, and FreeLB are all PGD-based adversarial training methods, which require multiple forward and back propagation iterations to calculate the optimal perturbation value. In the experiment, the number of forward-backward passes k is 3. Our method achieves state-of-the-art performance on five datasets by calculating the perturbation value through only one backpropagation. Note that we study the influence of the number of forward-backward propagation on DropAttack later. We can see that the performance of DropAttack-(I\&W) is better than DropAttack-(I) on all five datasets, which shows that the weight's adversarial training is work. In addition, based on the results in Table 2, our method has more improvements on the NLP datasets and less on the CV datasets, which is consistent with the experimental results in other adversarial training research papers \citep{cheng2019robust,zhao2018generating,zhu2020freelb}. We believe that the reason for this phenomenon is that the perturbation is directly added to the original pixel value for the picture, and the text is not directly modified to the word, the perturbation value is added to the embedding vector. The pixel value of the image is fixed, and the perturbation may change the distribution of the original sample. However, the word vector of the text is not unique and certain, so it is more likely to learn a better word vector after adding perturbation.

\begin{figure*}[]
	\centering
	\includegraphics[width=0.97\linewidth]{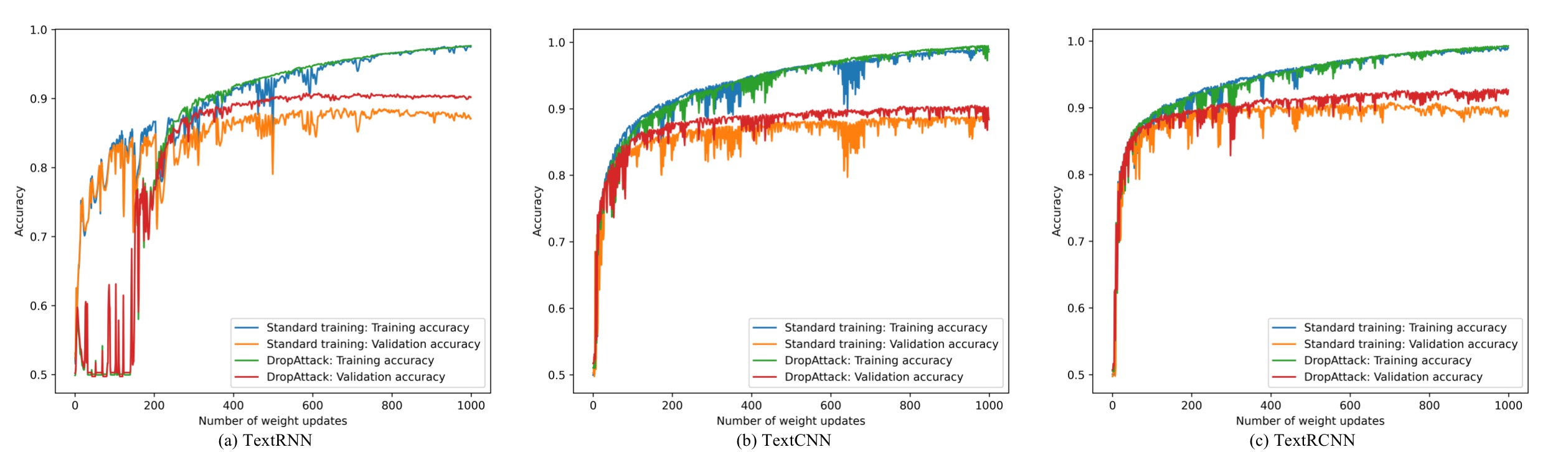}
	\caption{Training and validation accuracy of different models (TextRNN, TextCNN, RCNN) with and without DropAttack on IMDB dataset.
	}
	\label{f2}
\end{figure*}

In order to show the effect of DropAttack in preventing neural network overfitting more clearly, classification experiments were done with many different models of keeping all hyperparameters, including e and p, fixed. As shown in Figure \ref{f2}, Training and validation accuracy obtained for these models of different architectures (TextRNN, TextCNN and TextRCNN) as training progresses. The training accuracy under DropAttack training is basically the same as that under standard training, but the validation accuracy is higher, which proves that using the DropAttack adversarial training method can alleviate model overfitting. Furthermore, we can see that DropAttack adversarial training may be more difficult to converge in the early stage, because it is more difficult to optimize the target under attack, but after enough weight updates and learning, the validation accuracy of the model is relatively more stable. The key point is that DropAttack gives a obvious improvement across all nerual networks of different architectures, without using hyperparameters that were tuned specififically for each architecture. 

In addition, we also study the efficiency of DropAttack under training datasets of different sizes. We divide the IMDB training set into different sizes, and use the LSTM model with the same structure as the above, and the experimental results are shown in Table \ref{t3}. Compared with standard training, the performance of the model using the DropAttack training method on all seven training datasets of different sizes has been improved by more than 2\%. Furthermore, we found that DropAttack can achieve and even exceed the accuracy of standard training using only half of the training data. For example, DropAttack can achieve 83.88\% using 2500 training data, and the accuracy of using 5000 data based on standard training is 82.62\%.

\begin{table}[]
	\caption{The ability of our method DropAttack to prevent overfitting under different sizes of training data. The hyperparameters are set to $k=3,  \epsilon_x= \epsilon_\theta =5 , p_x = p_\theta = 0.7$.}
	\label{t3}
	\centering
	\begin{tabular}{l|c|c|c|c|c|c|c|c}
		\hline
		\multirow{2}{*}{Methods} & \multicolumn{7}{c}{Size of the training set} \\ \cline{2-9} 
		&100 & 500 & 1000 & 2500 & 5000 & 10000 & 20000 & 40000 \\ \hline \hline
		Standard Training &63.26& 74.26 & 78.30 & 81.14 & 82.62 & 84.92 & 85.42 & 88.12 \\
		DropAttack-3 Training &65.46& 76.34 & 80.70 & 83.88 & 85.22 & 87.02 & 88.86 & 90.42 \\ \hline
		Improvement $\uparrow$ &2.20& 2.08 & 2.40 & 2.66 & 2.60 & 2.10 & 3.42 & 2.30 \\ \hline
	\end{tabular}
\end{table}

\begin{table}[]
	\caption{Comparing the performance of DropAttack under different number of forward-backward propagation K. The reported results are calculated from 5 runs with the same hyper-parameters.}
	\label{t5}
	\centering
	\begin{tabular}{l|c|c|c|c|c}
		\hline
		Methods & IMDB & PHEME & AGnews & MNIST & CIFAR-10 \\ \hline \hline
		DropAttack-1 & 90.36 & 87.15/82.31 & 93.37 & 99.27 & 86.09 \\ \hline
		DropAttack-2 & 90.38 & 87.27/82.43 & 93.38 & 99.24 & 86.09 \\
		DropAttack-3 & 90.42 & 87.36/82.78 & 93.34 & 99.27 & 86.07 \\
		DropAttack-4 & 90.43 & 87.25/82.63 & 93.41 & 99.26 & 86.10 \\
		DropAttack-5 & 90.42 & 87.26/82.67 & 93.39 & 99.25 & 86.09 \\ \hline
	\end{tabular}
\end{table}

PGD-based DropAttack-k. we study the influence of the number of forward-backward propagation on DropAttack, and the experimental results are shown in Table \ref{t5}. We can find that multiple iterative calculations can indeed further improve the generalization of the neural network, because multiple iterations are more likely to get the optimal disturbance value. However, it is clear that more forward-backward propagation will greatly increase the training time. Therefore, a reasonable number of iterations K can be selected based on time and computing resources.

\section{Theoretical Analysis}
\label{5}
We provide another theoretical perspectives to explain why the adversarial training method DropAttack can be used as regularization to improve the generalization of the model and prevent overfitting. 
According to Section \ref{3}, the task of DropAttack can be to minimize the maximum internal adversarial risk, that is, to approximately optimize the following goals:
\begin{equation}\label{k8}
\mathop{{\rm min}}\limits_{\bm{\theta}} \mathbb{E}_{(x,y)\sim D}[\mathop{{\rm max}}\limits_{r_{x}\in S}L(\bm{\theta},\bm{x}+\bm{M_{x}}\cdot \bm{r_x},y)+\mathop{{\rm max}}\limits_{r_{{\theta}}\in S}L(\bm{\theta}+\bm{M_{\theta}}\cdot \bm{r_\theta},\bm{x},y) ]
\end{equation}
For formula \ref{k8}, we Taylor expand functions $f(x)=L(\bm{\theta},\bm{x}+\bm{M_{x}}\cdot \bm{r_x},y)$ and $f(\theta)=L(\bm{\theta}+\bm{M_{\theta}}\cdot \bm{r_\theta},\bm{x},y) $ at points $(\bm{x}+\bm{M_{x}}\cdot \bm{r_x})$ and $(\bm{\theta}+\bm{M_{\theta}}\cdot \bm{r_\theta})$ respectively:
\begin{equation}\label{k9}
\begin{split}
\mathop{{\rm min}}\limits_{\bm{\theta}} \mathbb{E}_{(x,y)\sim D}\{&\mathop{{\rm max}}\limits_{r_{x}\in S}[L(\bm{\theta},\bm{x},y)+< \nabla_{\bm{x}}L(\bm{\theta},\bm{x},y),\bm{M_{x}}\cdot \bm{r_x}>] \\ +&\mathop{{\rm max}}\limits_{r_{{\theta}}\in S}[L(\bm{\theta},\bm{x},y)+<\nabla_{\bm{{\theta}}}L(\bm{\theta},\bm{x},y),\bm{M_{\theta}}\cdot \bm{r_\theta}> ]\}
\end{split}
\end{equation}
Then, after substituting the values $\bm{r_x}= \epsilon_x \cdot \nabla_{\bm{x}}L(\bm{\theta},\bm{x},y) /\Vert\nabla_{\bm{x}}L(\bm{\theta},\bm{x},y)\Vert_2 ,~
\bm{r_{\theta}}= \epsilon_{\theta} \cdot \nabla_{\bm{{\theta}}}L(\bm{\theta},\bm{x},y) /\Vert\nabla_{\bm{{\theta}}}L(\bm{\theta},\bm{x},y)\Vert_2$   of the perturbation that maximize the antagonistic Loss, the following formula \ref{k10} is obtained:
\begin{equation}\label{k10}
\begin{split}
\mathop{{\rm min}}\limits_{\bm{\theta}} \mathbb{E}_{(x,y)\sim D}&[L(\bm{\theta},\bm{x},y)+< \nabla_{\bm{x}}L(\bm{\theta},\bm{x},y),\bm{M_{x}}\cdot \epsilon_x \cdot \nabla_{\bm{x}}L(\bm{\theta},\bm{x},y) /\Vert\nabla_{\bm{x}}L(\bm{\theta},\bm{x},y)\Vert_2> \\+ &L(\bm{\theta},\bm{x},y)+<\nabla_{\bm{{\theta}}}L(\bm{\theta},\bm{x},y),\bm{M_{\theta}}\cdot \epsilon_{\theta} \cdot \nabla_{\bm{{\theta}}}L(\bm{\theta},\bm{x},y) /\Vert\nabla_{\bm{{\theta}}}L(\bm{\theta},\bm{x},y)\Vert_2> ]
\end{split}
\end{equation}
\begin{equation}\label{key}
\begin{split}
\Rightarrow \mathop{{\rm min}}\limits_{\bm{\theta}} \mathbb{E}_{(x,y)\sim D}&[L(\bm{\theta},\bm{x},y)+ \epsilon_x \cdot\bm{M_{x}} < \nabla_{\bm{x}}L(\bm{\theta},\bm{x},y), \nabla_{\bm{x}}L(\bm{\theta},\bm{x},y) /\Vert\nabla_{\bm{x}}L(\bm{\theta},\bm{x},y)\Vert_2> \\+ L(\bm{\theta}&,\bm{x},y)+\epsilon_{\theta} \cdot\bm{M_{\theta}}<\nabla_{\bm{{\theta}}}L(\bm{\theta},\bm{x},y), \nabla_{\bm{{\theta}}}L(\bm{\theta},\bm{x},y) /\Vert\nabla_{\bm{{\theta}}}L(\bm{\theta},\bm{x},y)\Vert_2> ]
\end{split}
\end{equation}
\begin{equation}\label{key}
\begin{split}
\Rightarrow \mathop{{\rm min}}\limits_{\bm{\theta}} \mathbb{E}_{(x,y)\sim D}[2L(\bm{\theta},\bm{x},y)+ \epsilon_x \cdot  \Vert\bm{M_{x}}\cdot\nabla_{\bm{x}}L(\bm{\theta},\bm{x},y)\Vert_2+  \epsilon_{\theta} \cdot \Vert\bm{M_{\theta}}\cdot\nabla_{\bm{{\theta}}}L(\bm{\theta},\bm{x},y)\Vert_2 ]
\end{split}
\end{equation}
\begin{equation}\label{k13}
\begin{split}
\Rightarrow \mathop{{\rm min}}\limits_{\bm{\theta}}\mathbb{E}_{(x,y)\sim D} [2{\rm Loss}+ \epsilon_x \cdot  \Vert\bm{M_{x}}\cdot \bm{g_x}\Vert_2+  \epsilon_{\theta} \cdot \Vert\bm{M_{\theta}}\cdot\bm{g_{\theta}}\Vert_2 ]
\end{split}
\end{equation}

We can see the final optimization function formula \ref{k13}, which actually adds the implicit gradient regularization $\epsilon_x \cdot  \Vert\bm{M_{x}}\cdot \bm{g_x}\Vert_2$ and $\epsilon_{\theta} \cdot \Vert\bm{M_{\theta}}\cdot\bm{g_{\theta}}\Vert_2$ to a certain proportion of input $\bm{x}$ and parameters $\bm{\theta}$ after loss every time the parameter $\bm{\theta}$ is updated. 
Gradient penalty pushes the gradient of some parameters and inputs to approach zero, so that the model is likely to be optimized to a flatter minimum.

In order to further visually analyze the effectiveness of the proposed method, we draw the high-dimensional non-convex loss function with a visualization method proposed by \citep{li2018visualizing}. We visualize the loss landscapes around the minima of the empirical risk generated by standard training or DropAttack, the 2D visualization are plotted in Figure 3 and the 3D visualization are in Figure 4. Additional loss visualization are provided in the Appendix. Define two direction vectors, $\alpha$ and $\beta$ with the same dimensions as $\theta$, drawn from a Gaussian distribution with zero mean and a scale of the same order of magnitude as the variance of layer weights. Then we choose a center point $\theta^*$ and add a linear combination of $\alpha$ and $\beta$ to obtain a loss that is a function of the contribution of the two random direction vectors.
And we define a grid of points to evaluate the loss on i.e. range of values for $\delta$ and $\eta$ for which $L(\delta,\eta)$ is evaluated
and stored. 
\begin{equation*}
L(\delta,\eta) = \mathcal{L}(\theta^*+\delta \alpha + \delta \beta)
\end{equation*}
The results show that the test loss $L(\delta,\eta)$becomes lower and flatter during the
training with DropAttack. And DropAttack indeed selects flatter loss landscapes via masked adversarial perturbations. Many studies have shown that a flatter loss landscape usually means better generalization \citep{hochreiter1997long,keskar2019large,ishida2020we}.
\begin{figure*}[]
	\centering
	\includegraphics[width=0.93\linewidth]{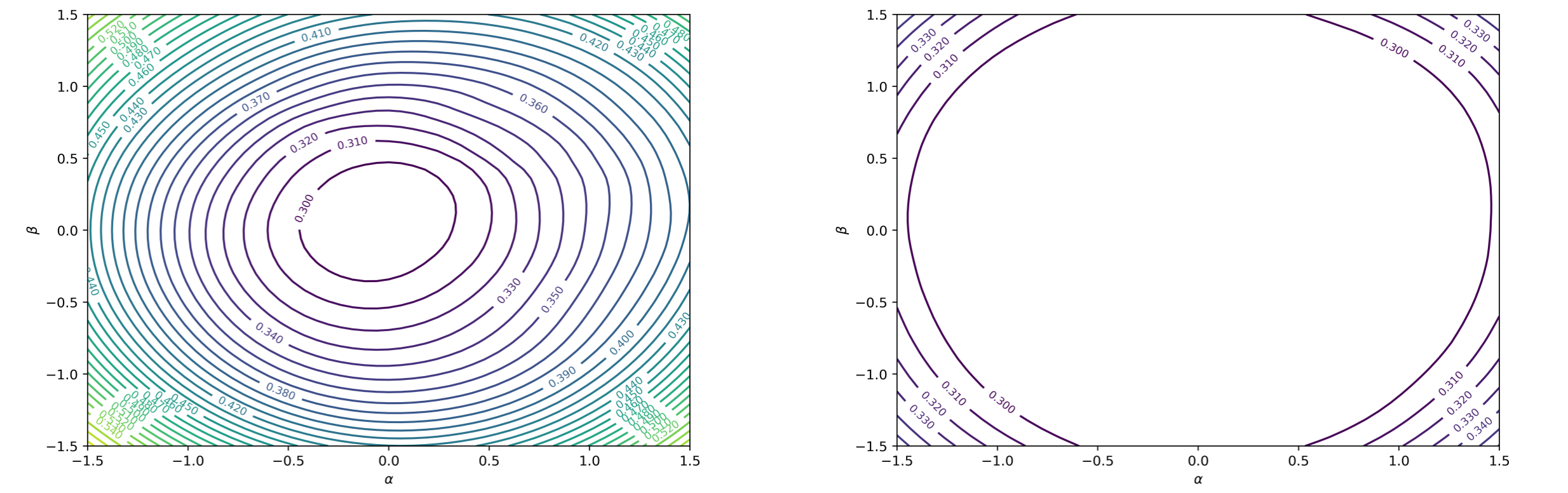}
	\caption{2D visualization of the minima of the empirical risk generated by standard training (left) and DropAttack (right) on IMDB dataset.
	}
	\label{f3}
\end{figure*}
\begin{figure*}[]
	\centering
	\includegraphics[width=0.95\linewidth]{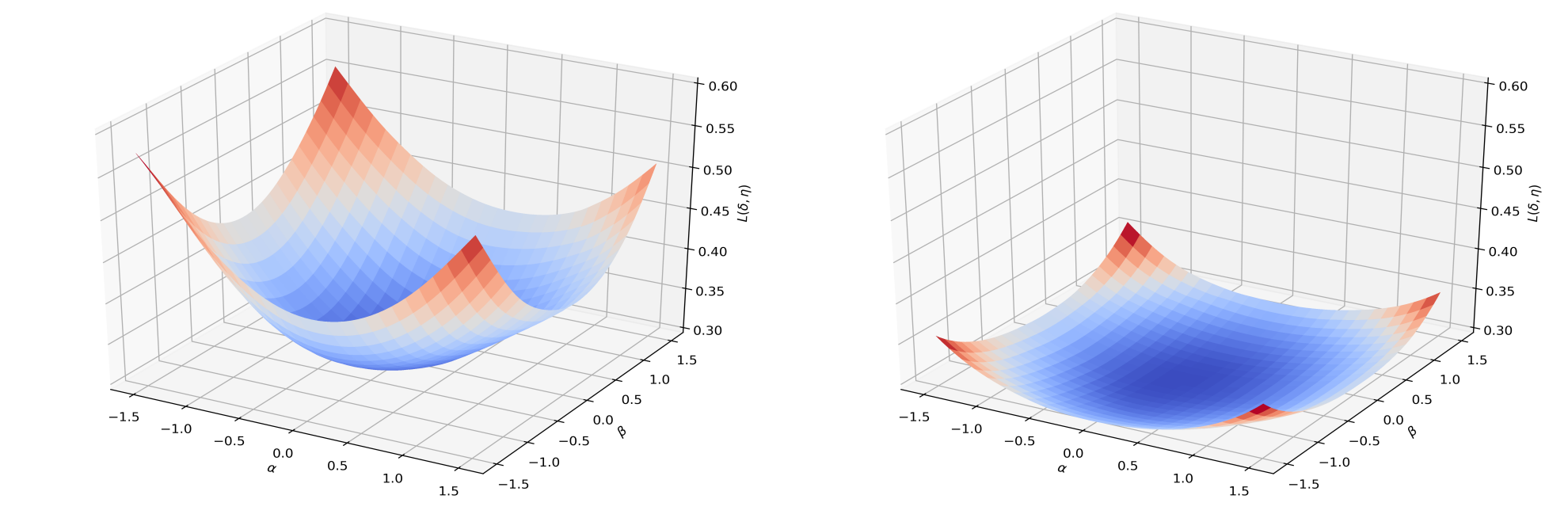}
	\caption{3D visualization of the minima of the empirical risk generated by standard training (left) and DropAttack (right) on IMDB dataset.
	}
	\label{f4}
\end{figure*}
\section{Conclusion}
In this work, we propose a masked weight adversarial training method, DropAttack, to improve the generalization ability of neural network models and prevent overfitting. The proposed algorithm uses the gradient method to attack the input and weight parameters according to a certain probability, and enhances the generalization of the model by minimizing the resultant adversarial risk. Experimental results prove that DropAttack can effectively improve the generalization of models and prevent overfitting, especially in the field of NLP. In addition, we theoretically proved that our algorithm can regularize the gradient of model parameters. Therefore, DropAttack can improve the robustness and generalization of the model. The current adversarial training still consumes more computing resources and time than the standard stochastic gradient descent, so it is a valuable research direction to accelerate the adversarial training while improving generalization in the future.


\bibliography{iclr2022_conference}
\bibliographystyle{iclr2022_conference}

\newpage
\appendix
\section{Additional Experimental Details and Results }
\label{a}
In order to prove the effectiveness of DropAttack, we conducted many experiments on five datasets: IMDB, PHEME, AGnews, MNIST, CIFAR-10. The detailed experimental settings and results are shown in Table 6, Table 7, Table 8, Table 9, Table 10. The square brackets after DropAttack indicate the object of perturbation. For example, DropAttack[Embedding, Lstm.ih.w] means adding perturbation to Embedding and Lstm.ih.w parameters. 

\begin{table}[h]
	\centering
	\caption{Experimental details on the IMDB dataset.}
	\label{ti}
	\begin{tabular}{l|c}
		\hline
		Method & Accuracy (\%) \\ \hline \hline
		LSTM & 88.12 \\ \hline
		LSTM + DropAttack{[}Embedding{]} ( e = 5, p = 0.5 ) & 89.76 \\ \hline
		LSTM + DropAttack{[}Embedding, Fc.w{]} ( e = 5, p = 0.5 ) & 88.60 \\ \hline
		LSTM + DropAttack{[}Lstm.hh.w{]} ( e = 5, p = 0.5 ) & 86.64 \\ \hline
		LSTM + DropAttack{[}Lstm.ih.w{]} ( e = 5, p = 0.5 ) & 89.80 \\ \hline
		LSTM + DropAttack{[}Embedding, Lstm.hh.w{]} ( e = 5, p = 0.5 ) & 87.90 \\ \hline
		LSTM + DropAttack{[}Embedding, Lstm.ih.w{]} ( e = 5, p = 0.5 ) & 90.21 \\ \hline
		LSTM + DropAttack{[}Embedding, Lstm.ih.w{]} ( e = 3, p = 0.7 ) & 90.22\\ \hline
		LSTM + DropAttack{[}Embedding, Lstm.ih.w{]} ( e = 5, p = 0.7 ) & 90.34\\ \hline
		LSTM + DropAttack{[}Embedding, Lstm.ih.w{]} ( e = 7, p = 0.7 ) & \textbf{90.36} \\ \hline
		LSTM + DropAttack{[}Embedding, Lstm.hh.w, Lstm.ih.w{]} ( e = 5, p = 0.5 ) & 89.56 \\ \hline
		LSTM + DropAttack{[}Embedding, Lstm.hh.w, Lstm.ih.w{]} ( e = 5, p = 0.6 ) & 89.57 \\ \hline
	\end{tabular}
\end{table}

\begin{table}[h]
	\centering
	\caption{Experimental details on the PHEME dataset.}
	\label{tp}
	\begin{tabular}{l|c}
		\hline
		Method & Accuracy/F1 score (\%) \\ \hline 	\hline
		BiLSTM & 84.08/78.99 \\ \hline
		BiLSTM + DropAttack{[}Embedding{]} ( e = 5, p = 0.5 ) & 85.69/79.97 \\ \hline
		BiLSTM + DropAttack{[}Lstm.hh.w{]} ( e = 5, p = 0.5 ) & 86.00/80.03 \\ \hline
		BiLSTM + DropAttack{[}Lstm.ih.w{]} ( e = 5, p = 0.5 ) & 83.40/78.64 \\ \hline
		BiLSTM + DropAttack{[}Fc.w{]} ( e = 5, p = 0.5 ) & 84.60/79.32 \\ \hline
		BiLSTM + DropAttack{[}Embedding, Lstm.hh.w{]} ( e = 5, p = 0.5 ) & 87.14/81.02 \\ \hline
		BiLSTM + DropAttack{[}Embedding, Lstm.hh.w{]} ( e = 5, p = 0.6 ) & 87.14/81.04 \\ \hline
		BiLSTM + DropAttack{[}Embedding, Lstm.ih.w{]} ( e = 5, p = 0.5 ) & 86.74/80.13 \\ \hline
		BiLSTM + DropAttack{[}Embedding, Lstm.hh.w{]} ( e = 5, p = 0.7 ) & \textbf{87.15/81.31} \\ \hline
		BiLSTM + DropAttack{[}Embedding, Lstm.hh.w{]} ( e = 5, p = 0.8 ) & 87.11/81.24 \\ \hline
		BiLSTM + DropAttack{[}Embedding, Lstm.hh.w, Lstm.ih.w{]} ( e = 5, p = 0.5 ) & 85.54/79.57 \\ \hline
		BiLSTM + DropAttack{[}Embedding, Lstm.hh.w, Lstm.ih.w{]} ( e = 5, p = 0.7 ) & 85.35/79.36 \\ \hline
	\end{tabular}
\end{table}

\begin{table}[h]
	\centering
	\caption{Experimental details on the AGnews dataset.}
	\label{ta}
	\begin{tabular}{l|c}
		\hline
		Method & Accuracy (\%) \\ \hline \hline
		BiGRU & 91.87 \\ \hline
		BiGRU+ DropAttack{[}Embedding{]} ( e = 5, p = 0.5 ) & 93.35 \\ \hline
		BiGRU+ DropAttack{[}Embedding{]} ( e = 5, p = 0.7 ) & 93.34 \\ \hline
		BiGRU + DropAttack{[}Gru.hh.w{]} ( e = 5, p = 0.5 ) & 92.25 \\ \hline
		BiGRU + DropAttack{[}Gru.ih.w{]} ( e = 5, p = 0.5 ) & 92.46 \\ \hline
		BiGRU + DropAttack{[}Gru.hh.w, Gru.ih.w{]} ( e = 5, p = 0.5 ) & 92.70 \\ \hline
		BiGRU + DropAttack{[}Fc.w{]} ( e = 5, p = 0.5 ) & 92.24 \\ \hline
		BiGRU + DropAttack{[}Embedding, Gru.ih.w{]} ( e = 5, p = 0.5 ) & 93.12 \\ \hline
		BiGRU + DropAttack{[}Embedding, Gru.ih.w{]} ( e = 5, p = 0.7 ) & \textbf{93.37} \\ \hline
		BiGRU + DropAttack{[}Embedding, Gru.hh.w{]} ( e = 5, p = 0.5 ) & 92.70 \\ \hline
		BiGRU + DropAttack{[}Embedding, Gru.hh.w, Gru.ih.w{]} ( e = 5, p = 0.5 ) & 93.12 \\ \hline
		BiGRU + DropAttack{[}Embedding, Gru.ih.w, Gru.ih.w.reverse{]} ( e = 5, p = 0.5 ) & 92.88 \\ \hline
	\end{tabular}
\end{table}

\begin{table}[h]
	\centering
	\caption{Experimental details on the MNIST dataset.}
	\label{tm}
	\begin{tabular}{l|c}
		\hline
		Method & Accuracy (\%) \\ \hline  \hline
		LeNet-5 & 98.95 \\ \hline
		LeNet-5  + DropAttack{[}Input{]} ( e = 5, p = 0.5 ) & 99.16 \\ \hline
		LeNet-5  + DropAttack{[}Conv.1.w{]} ( e = 5, p = 0.5 ) & 99.08 \\ \hline
		LeNet-5  + DropAttack{[}Input, Conv.1.w{]} ( e = 5, p = 0.5 ) & \textbf{99.27} \\ \hline
		LeNet-5  + DropAttack{[}Input, Conv.1.w{]} ( e = 5, p = 0.7 ) & 99.25 \\ \hline
		LeNet-5  + DropAttack{[}Input, Conv.2.w{]} ( e = 5, p = 0.5 ) & 99.12 \\ \hline
		LeNet-5  + DropAttack{[}Input, Conv.1.b{]} ( e = 5, p = 0.5 ) & 99.10 \\ \hline
		LeNet-5  + DropAttack{[}Conv.2.w{]} ( e = 5, p = 0.5 ) & 99.11 \\ \hline
		LeNet-5  + DropAttack{[}Conv.1.b, Conv.2.w{]} ( e = 5, p = 0.5 ) & 99.09 \\ \hline
		LeNet-5  + DropAttack{[}Conv.1.w, Conv.2.w{]} ( e = 5, p = 0.5 ) & 98.78 \\ \hline
		LeNet-5  + DropAttack{[}Conv.1.w, Fc.1{]} ( e = 5, p = 0.5 ) & 98.93 \\ \hline
		LeNet-5  + DropAttack{[}Conv.2.w, Fc.1{]} ( e = 5, p = 0.5 ) & 99.10 \\ \hline
		LeNet-5  + DropAttack{[}Conv.2.w, Fc.2{]} ( e = 5, p = 0.5 ) & 99.05 \\ \hline
		LeNet-5  + DropAttack{[}Conv1.w, Conv2.w, fc1.w2{]} ( e = 5, p = 0.5 ) & 98.48 \\ \hline
	\end{tabular}
\end{table}

\begin{table}[]
	\centering
	\caption{Experimental details on the CIFAR-10 dataset.}
	\label{tc}
	\begin{tabular}{l|c}
		\hline
		Method & Accuracy (\%) \\ \hline  \hline
		VGGNet-16 & 84.67 \\ \hline
		VGGNet-16 + DropAttack{[}Input{]} ( e = 5, p = 0.5 ) & 86.02 \\ \hline
		VGGNet-16 + DropAttack{[}Conv.1.w{]} ( e = 5, p = 0.5 ) & 83.61 \\ \hline
		VGGNet-16 + DropAttack{[}Input, Conv.1.w{]} ( e = 5, p = 0.5 ) & \textbf{86.09} \\ \hline
		VGGNet-16 + DropAttack{[}Input, Conv.1.w{]} ( e = 5, p = 0.7 ) & 86.02 \\ \hline
		VGGNet-16 + DropAttack{[}Input, Conv.3.w{]} ( e = 5, p = 0.7 ) & 85.13 \\ \hline
		VGGNet-16 + DropAttack{[}Input, Conv.5.w{]} ( e = 5, p = 0.7 ) & 85.13 \\ \hline
		VGGNet-16 + DropAttack{[}BatchNorm.1.w{]} ( e = 5, p = 0.5 ) & 85.51 \\ \hline
		VGGNet-16 + DropAttack{[}Conv.2.w{]} ( e = 5, p = 0.5 ) & 83.16 \\ \hline
		VGGNet-16 + DropAttack{[}Conv.6.w{]} ( e = 5, p = 0.5 ) & 85.27 \\ \hline
		VGGNet-16 + DropAttack{[}BatchNorm.8.w{]} ( e = 5, p = 0.5 ) & 85.32 \\ \hline
		VGGNet-16 + DropAttack{[}Conv.1.w, BatchNorm.1.w{]} ( e = 5, p = 0.5 ) & 84.41 \\ \hline
		VGGNet-16 + DropAttack{[}Input, Conv.1.w, BatchNorm.1.w{]} ( e = 5, p = 0.5 )  & 85.01 \\ \hline
	\end{tabular}
\end{table}
For an important discussion and research question: In addition to the perturbation of the input layer, which layer's weight parameters are better to perturb? According to our experimental results and experience, for the perturbation of hidden layer parameters, the effect of being close to the input layer will be better than that of being close to the output layer. Essentially, the perturbation of the hidden layer is the perturbation of the higher-dimensional embedding of the input. Due to the highly linearization of neural networks, small changes in the input vector may result in changes in the output of multiple layers. Therefore, the deeper weight parameters need to be robust enough to resist overfitting and prevent fitting to overly sensitive input features. 
\section{Hyperparameter sensitivity analysis}
\begin{table}[h]
	\caption{The influence of hyperparameters perturbation coefficient e and attack probability p on model performance. The tested dataset is IMDB, and the model structure is the same as that in Table \ref{t2}. The top ten performance is emphasized in \textbf{bold}.}
	\label{t4}
	\centering
	\begin{threeparttable}
		\begin{tabular}{c|c|c|c|c|c|c|c|c}
			\hline
			\multicolumn{2}{c}{} & \multicolumn{7}{c}{Perturbation coefficient \tnote{*}} \\ \cline{3-9} 
			\multicolumn{2}{c}{} & $\epsilon$ = 0.01 & $\epsilon$ = 0.1 & $\epsilon$ = 1 & $\epsilon$ = 3 & $\epsilon$ = 5 & $\epsilon$ = 7 & $\epsilon$ = 9 \\ \hline \hline
			\multirow{7}{*}{\begin{tabular}[c]{@{}c@{}}Attack\\ probability\tnote{*}\end{tabular}} & \multicolumn{1}{c|}{\begin{tabular}[c]{@{}c@{}}P = 0.0\end{tabular}} & 88.12 & 88.12 & 88.12 & 88.12 & 88.12 & 88.12 & 88.12 \\
			& \multicolumn{1}{c|}{P = 0.1} & 89.60 & 89.78 & 89.40 & 88.74 & 89.30 & 88.88 & 89.26 \\
			& \multicolumn{1}{c|}{P = 0.3} & 89.98 & 90.12 & 90.02 & 90.10 & 90.04 & \textbf{90.28} & 89.74 \\
			& \multicolumn{1}{c|}{P = 0.5} & 90.13 & 90.16 & 90.01 & \textbf{90.25} & \textbf{90.21} & \textbf{90.30} & 89.94 \\
			& \multicolumn{1}{c|}{P = 0.7} & 90.17 & \textbf{90.32} & \textbf{90.18} & \textbf{90.20} & \textbf{90.18} & \textbf{90.36} & 90.14 \\
			& \multicolumn{1}{c|}{P = 0.9} & 89.76 & \textbf{90.22} & 90.16 & 89.22 & 90.12 & 90.04 & 90.18 \\
			& \multicolumn{1}{c|}{\begin{tabular}[c]{@{}c@{}}P = 1.0\end{tabular}} & 89.86 & 89.54 & 89.74 & 89.40 & 90.02 & 89.90 & 90.10 \\ \hline
		\end{tabular}
		\begin{tablenotes}
			\footnotesize
			\item[*] {Note that $\epsilon_x$ and $\epsilon_\theta$ are uniformly denoted by $\epsilon$; $p_x$ and $p_\theta$ are uniformly denoted by $p$.} 
		\end{tablenotes}
	\end{threeparttable}
\end{table}

DropAttack has three tunable hyperparameter perturbation coefficient $\epsilon$, attack probability p (the probability of attacking a weight param in the network) and number of forward-backward propagation K. We explore the effect of varying these hyperparameter. Firstly, We fixed the value of K to 1, and let $\epsilon$ take the values in [0.001, 0.1, 1, 3, 5, 7, 9] in turn, and p take the values in [0, 0.1, 0.3, 0.5, 0.7, 0.9, 1] in turn, so that there is a total of 7 x 7 = 49 kinds of hyperparameter combinations, and the experimental results are shown in Table \ref{t4}. The best performance is 90.36\% when $ \epsilon = 7$ and p = 0.7. When p = 0, it means that the model is standard training without any attack, and its accuracy is the lowest. And we found that the effect is significantly worse when p is less than 0.3 or greater than 0.9. When p =1, random masking is not used, although the performance is improved but not much, because the attack combination is relatively single. Through experimental results, we think that the hyperparameter p is best to be between 0.5 and 0.7. 

\begin{figure*}[h]
	\centering
	\includegraphics[width=0.99\linewidth]{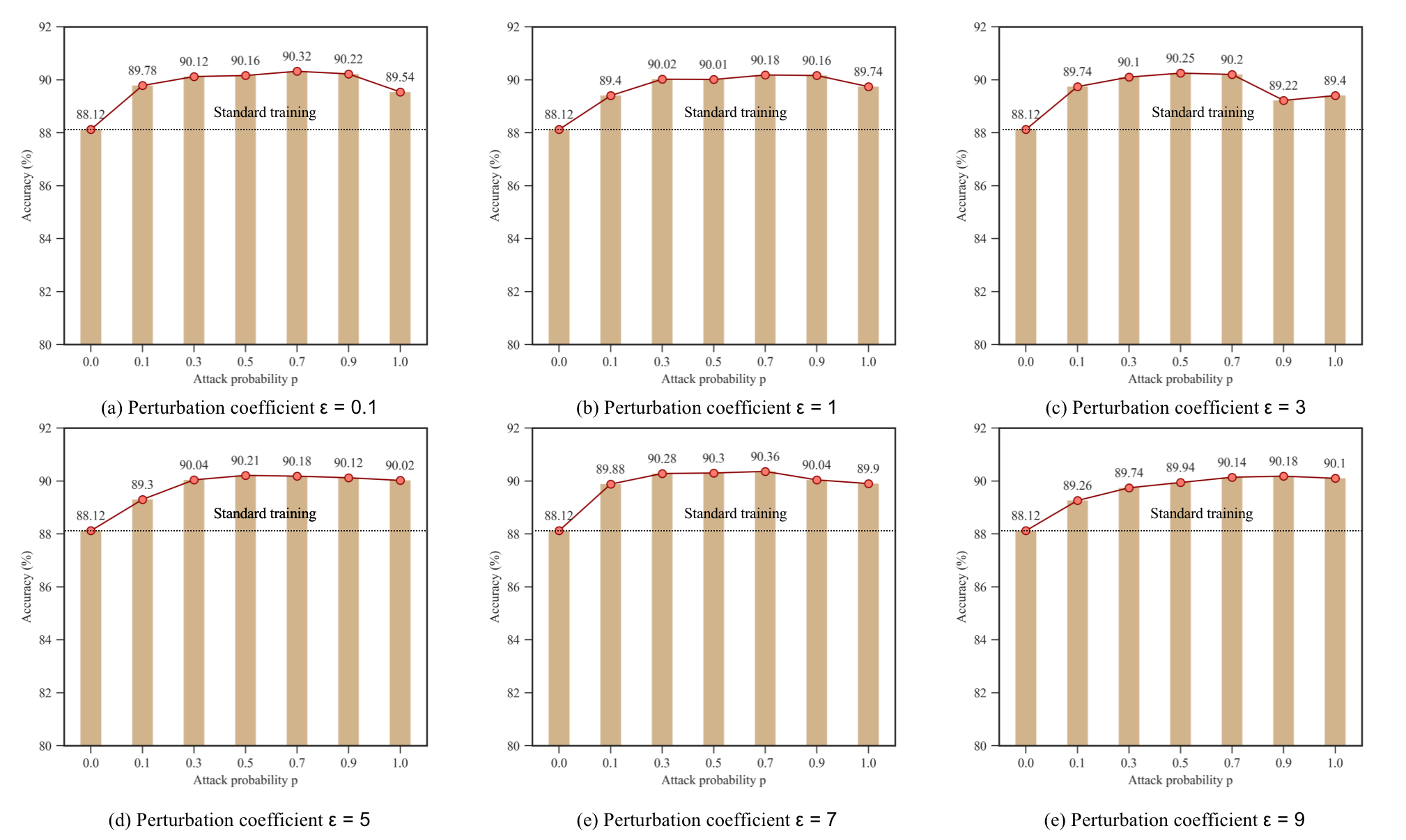}
	\caption{The impact of attack probability ($p\in [0, 0.1, 0.3, 0.5, 0.7, 0.9, 1]$) on model performance under a fixed perturbation coefficient coefficient ($\epsilon \in [ 0.1, 1, 3, 5, 7, 9] $).
	}
	\label{f5}
\end{figure*}

As shown in Figure \ref{f5}, we study the impact of different attack probabilities on model performance under different perturbation coefficient coefficients. We can see that the attack probability ranges from 0 to 0.7, and the performance of the model increases as the attack probability increases, because the intensity of the model attack is getting stronger. However, we found that the attack probability increased from 0.7 to 1, and our performance decreased instead. Because an excessively high attack probability will reduce the diversity of attack combinations, when p=0, it is approximately a standard adversarial attack.
\newpage

\section{Additional Loss Visualization}
\begin{figure*}[h]
	\centering
	\includegraphics[width=0.99\linewidth]{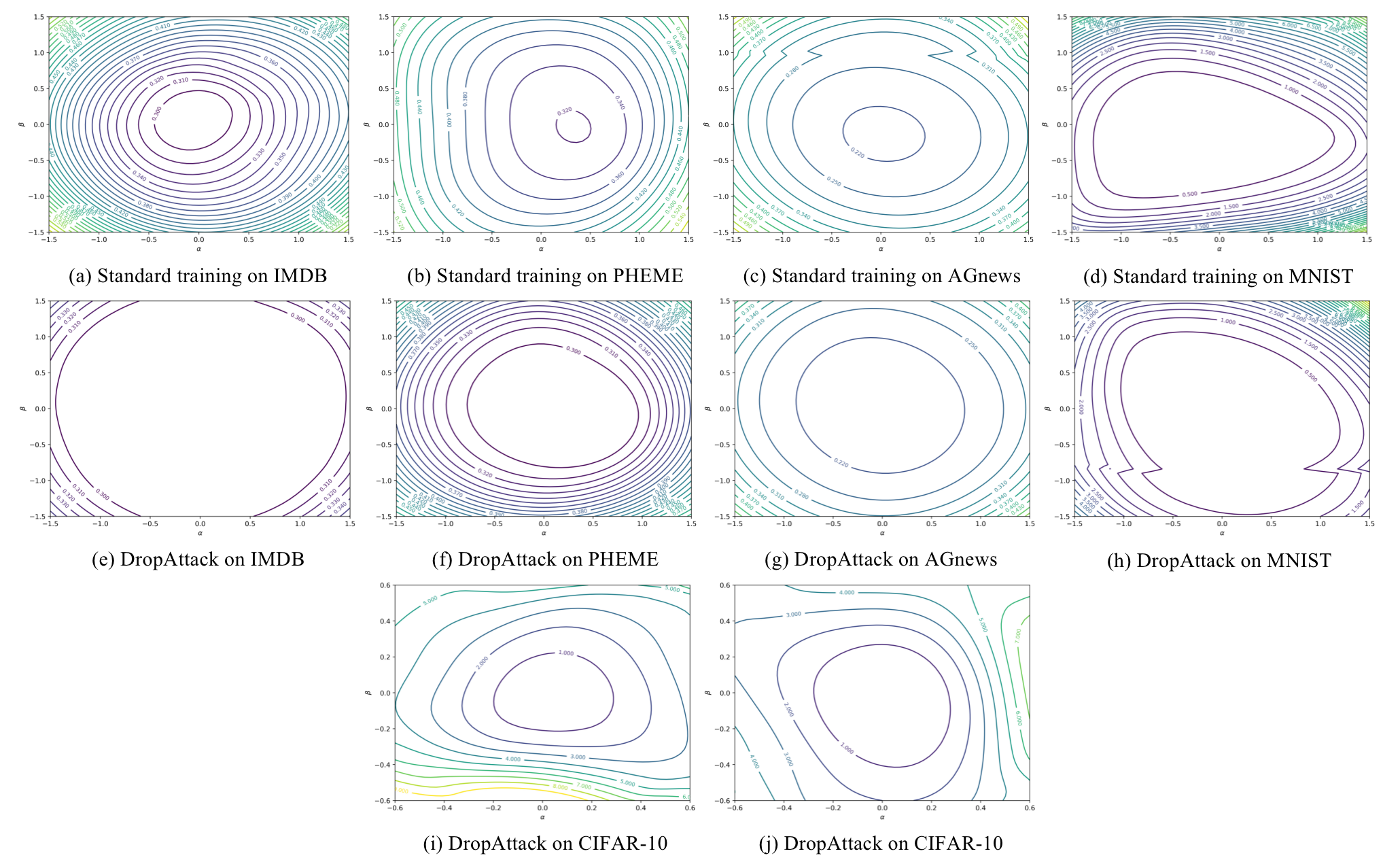}
	\caption{2D visualization of the minima of the empirical risk selected by standard training and DropAttack on IMDB, PHEME, AGnews, MNIST, CIFAR-10 datasets, respectively.
	}
	\label{f6}
\end{figure*}

\begin{figure*}[h]
\centering
\includegraphics[width=0.99\linewidth]{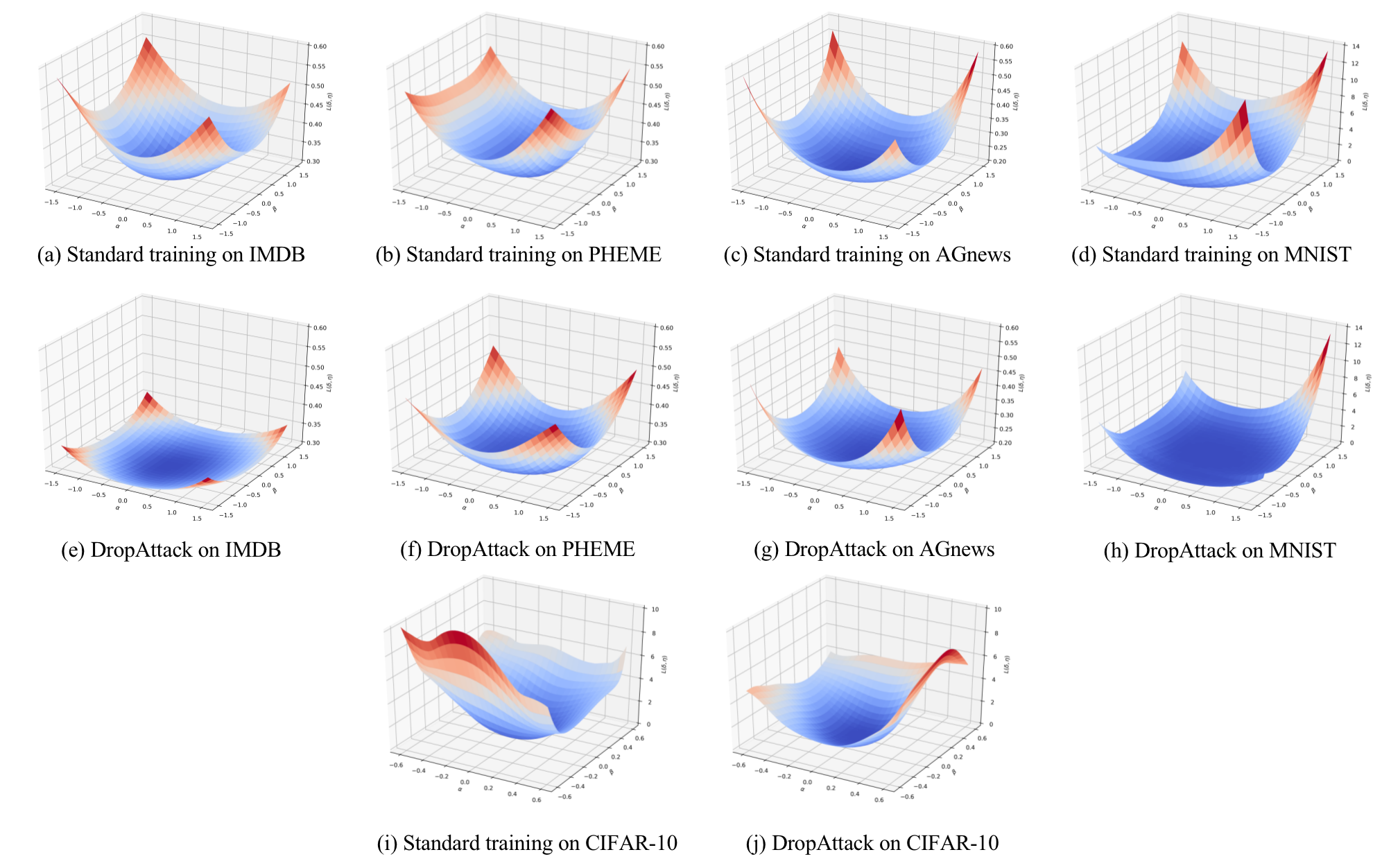}
\caption{3D visualization of the minima of the empirical risk selected by standard training and DropAttack on IMDB, PHEME, AGnews, MNIST, CIFAR-10 datasets, respectively.
}
\label{f7}
\end{figure*}

We visualize the test loss function landscapes of the standard training and DropAttack adversarial training models separately. The 2D and 3D visualization results are shown in Figure \ref{f6} and Figure \ref{f7}, respectively. The structure and parameters of the models are derived from Section 4.2.

\end{document}

%% file: math_commands.tex
\usepackage{amsmath,amsfonts,bm}

\def\eqref#1{equation~\ref{#1}}

\def\1{\bm{1}}

\DeclareMathAlphabet{\mathsfit}{\encodingdefault}{\sfdefault}{m}{sl}
\SetMathAlphabet{\mathsfit}{bold}{\encodingdefault}{\sfdefault}{bx}{n}

